\documentclass[10pt,twocolumn,letterpaper]{article}

\usepackage{cvpr}
\usepackage{times}
\usepackage{epsfig}
\usepackage{graphicx}
\usepackage{amsmath}
\usepackage{amssymb}
\usepackage{subcaption}
\usepackage{booktabs}
\usepackage{soul}

\usepackage[pagebackref=true,breaklinks=true,letterpaper=true,colorlinks,bookmarks=false]{hyperref}

\cvprfinalcopy 

\def\cvprPaperID{****} 

\ifcvprfinal\fi
\begin{document}

\title{Light-weight Head Pose Invariant Gaze Tracking}

\author{Rajeev Ranjan\\
University of Maryland\\
{\tt\small rranjan1@umiacs.umd.edu}
\and
Shalini De Mello\\
NVIDIA\\
{\tt\small shalinig@nvidia.com}
\and
Jan Kautz\\
NVIDIA\\
{\tt\small jkautz@nvidia.com}
}

\maketitle


\begin{abstract}
Unconstrained remote gaze tracking using off-the-shelf cameras is a challenging problem.
Recently, promising algorithms for appearance-based gaze estimation using convolutional neural networks (CNN) have been proposed.
Improving their robustness to various confounding factors including variable head pose, subject identity, illumination and image quality remain open problems.
In this work, we study the effect of variable head pose on machine learning regressors trained to estimate gaze direction.
We propose a novel branched CNN architecture that improves the robustness of gaze classifiers to variable head pose, without increasing computational cost.
We also present various procedures to effectively train our gaze network including transfer learning from the more closely related task of object viewpoint estimation and from a large high-fidelity synthetic gaze dataset, which enable our ten times faster gaze network to achieve competitive accuracy to its current state-of-the-art direct competitor.
\end{abstract}


\section{Introduction}
Gaze tracking has numerous applications in the design of industrial and web interfaces: for enhancing automotive safety by monitoring the visual behavior of drivers, for increased accessibility via gaze-controlled interfaces, for correcting gaze direction during video conferencing, and for rendering foveated displays in the bandwidth-constrained environments of augmented and virtual reality headsets. \par
Unconstrained gaze tracking refers to calibration-free, subject-, viewpoint-, and illumination-independent gaze tracking using a remotely placed off-the-shelf camera. 
It allows for free head motion of the subject.
Recently, various promising appearance-based techniques for unconstrained remote gaze estimation have been proposed~\cite{sugano2014learning, zhang2015appearance, huang2015tabletgaze, krafka2016eye, zhang2017s, zhang2017mpiigaze}.
Among them, techniques based on convolutional neural networks (CNN) are the most successful~\cite{krafka2016eye, zhang2017mpiigaze}.
Researchers have also proposed computer graphics models to generate synthetic training data for gaze estimation~\cite{wood2015rendering, wood2016learning}, along with techniques to adapt them to real-world data~\cite{wood2015rendering, shrivastava2017learning}.
Nevertheless, many open problems remain to be solved in order to increase the accuracy of gaze tracking including robustness to variability in head pose, illumination, subject identity, image quality and distribution of head pose and gaze angles~\cite{zhang2017mpiigaze}.
Furthermore, many applications of gaze tracking require operation in mobile environments on low-cost processors. 
Applications like foveated rendering require extremely fast gaze estimation (several hundred hertz)~\cite{patney2016perceptually}.
Hence, achieving realtime performance, without lowering quality is a practical concern for the deployment of gaze estimation technology in the real-world.
To improve computational efficiency, Krafkra \etal~\cite{krafka2016eye} use dark knowledge to reduce the size of their gaze CNN, but incur some loss of accuracy.

In this work we address several challenges of unconstrained gaze tracking.
First, we systematically study the effect of variable head pose on machine learning-based gaze regressors.
We show that one effect of variable head pose is the change in the prior distribution of gaze angles that the regressor needs to predict. 
This motivates the need for higher specificity in the design of gaze networks that process eye images captured from widely varying viewpoints.
To address this, first, we propose a novel and efficient head-pose-dependent branched CNN architecture to improve robustness to variable head pose, which does not increase the computational cost of inference.
Second, for realtime high-quality gaze estimation, we present various effective training procedures for gaze CNNs, which help our ten times faster CNN to achieve the same accuracy as the current state-of-the-art more complex algorithm~\cite{zhang2017mpiigaze}.
Our contributions to the training procedure include (a) transfer-learning from the closely related task of general object view-point estimation~\cite{su2015render} instead of from object classification -- the latter is the de-facto approach for gaze estimation~\cite{krafka2016eye, zhang2017mpiigaze}; and (b) from on a CNN trained on 1M synthetic photo-realistic images generated from the SynthesEyes model~\cite{wood2015rendering}.

\section{Related Work}


Research on conventional gaze trackers is summarized in~\cite{hansen2010eye}.
Recently, various appearance-based gaze tracking algorithms have been proposed that train machine learning regression functions using adaptive linear models~\cite{lu2014adaptive, mora2012gaze}, support vector machines~\cite{chuang2014estimating}, random forests (RF)~\cite{sugano2014learning, huang2015tabletgaze}, and more recently CNNs~\cite{zhang2015appearance, wood2015rendering, krafka2016eye, zhang2017s, zhang2017mpiigaze, deng2017monocular, shrivastava2017learning}. 
Other approaches use fitting to generative face and/or eye region models~\cite{alberto2014geometric, wood20163d, wang2017real}.
Among appearance-based approaches, CNNs~\cite{zhang2017s, zhang2017mpiigaze, deng2017monocular} are the most successful.

Inputs to gaze CNNs, investigated previously, include eye images and head orientation information~\cite{zhang2015appearance, wood2015rendering, zhang2017mpiigaze}, or full face images along with their location information~\cite{krafka2016eye, zhang2017s, deng2017monocular}.
Full-face-based CNNs are more accurate for gaze estimation than eye image-based CNNs~\cite{krafka2016eye, zhang2017s}, but are limited in their applicability as the entire face may not be visible to the camera in all situations~\cite{huang2015tabletgaze}.
Our algorithm uses information from the eye region only along with head orientation, and thus is more widely applicable.

Most existing algorithms regress 3D gaze angles, but a few directly regress the point-of-regard in 2D screen co-ordinates~\cite{krafka2016eye, zhang2017s}.
As Zhang \etal~\cite{zhang2017s} also note, computing 3D gaze directions is a more general solution, which, along with the 3D positions of the eye centers, can be used to compute any 3D point-of-regard.
Hence, we too compute the more general 3D gaze angles.

Many open challenges in gaze estimation including robustness to variable head pose, image resolution and quality, individual eye-shape, illumination and the presence of eye glasses are noted in~\cite{zhang2017mpiigaze}.
To deal with head pose, the early work~\cite{lu2014adaptive} limits head pose to close to frontal. 
Others generate frontal images of the face by capturing depth information in addition to color and warping~\cite{mora2012gaze, alberto2014geometric}.
Sugano \etal~\cite{sugano2014learning} create overlapping clusters of head pose and train one RF regressor per cluster.
At inference time, for each input, the outputs of multiple neighboring pose regressors are computed and averaged together, which improves accuracy, but because it involves inferencing via multiple regressors it also increases the computational cost.
Zhang \etal apply Sugano \etal's~\cite{sugano2014learning} approach of head pose clustering to two different gaze CNN architectures~\cite{zhang2015appearance, zhang2017mpiigaze} and observe a reduction in accuracy with it. We also use the idea of clustering head pose for gaze estimation. However, unlike the previous approaches, we employ late branching of one CNN, where only the few final fully-connected layers are specific to individual pose clusters and the rest of the layers are shared. We show that when large head poses are present this architecture results in improved accuracy without incurring any additional computational cost for inference. 

Using progressively more complex CNNs architectures (LeNet~\cite{lecun1995learning}, AlexNet~\cite{krizhevsky2012imagenet} and VGG~\cite{simonyan2014very}) and their weights trained on ImageNet as the bases for gaze networks has helped to improve the accuracy of gaze estimation~\cite{zhang2015appearance, zhang2017s, zhang2017mpiigaze}.
However, this comes at the expense of increased computational cost, which is a real challenge for deploying gaze CNNs on embedded devices.
Krafka \etal~\cite{krafka2016eye} explore reducing the size of their CNN using dark knowledge~\cite{hinton2015distilling}, but it results in reduction in accuracy of their model.
Orthogonal to these works, we propose effective training procedures including transfer learning from the closely related task of object view-point estimation and from a large photo-realistic synthetic dataset to train a much ($10\times$) faster and smaller CNN than Zhang \etal's~\cite{zhang2017mpiigaze} state-of-the-art network and still achieve competitive accuracy to it.

Lastly, there exist multiple synthetic models to generate synthetic training images for gaze CNNs: (a) UT Multi-view~\cite{sugano2014learning} with limited discrete gaze angles, (b) SynthesEyes~\cite{wood2015rendering} with 10 real head scans and highly photo-realistic rendering, and (c) UnityEyes~\cite{wood2016learning} with realtime rendering of less photo-realistic images, but with a morphable model to generate infinite shapes and textures for the head region surrounding the eyeball. Of these, large datasets (of a few millions images) generated with the UnityEyes model are used most widely ~\cite{wood2016learning, shrivastava2017learning}. 
A few works use a smaller dataset of 12K photo-realistic images generated from the SynthesEyes model~\cite{wood2015rendering, wood2016learning, zhang2017mpiigaze} and found it to be inferior to the 1M UnityEyes dataset in improving gaze estimation accuracy. Unlike these previous works, we generate a large dataset of 1M photo-realistic images using the SynthesEyes model and show its effectiveness in improving the accuracy of gaze estimation.


\begin{figure*}[t]
\centering
\begin{tabular}{ccc}
\includegraphics[scale=0.5]{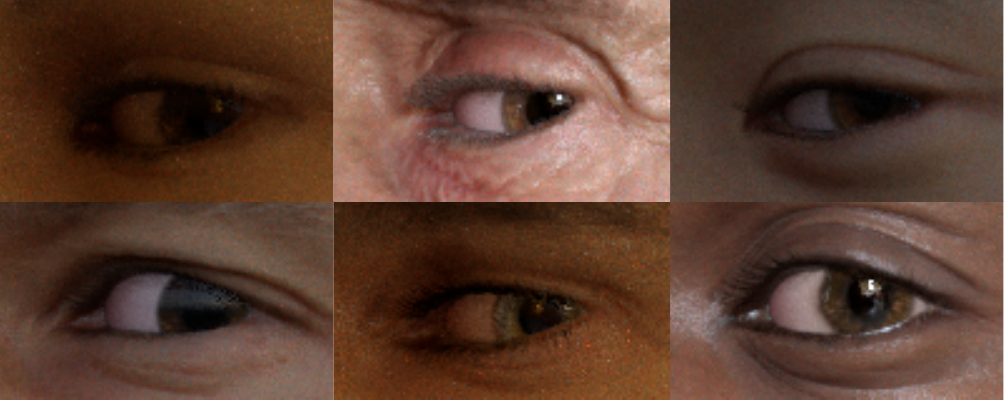}
&
\includegraphics[scale=0.5]{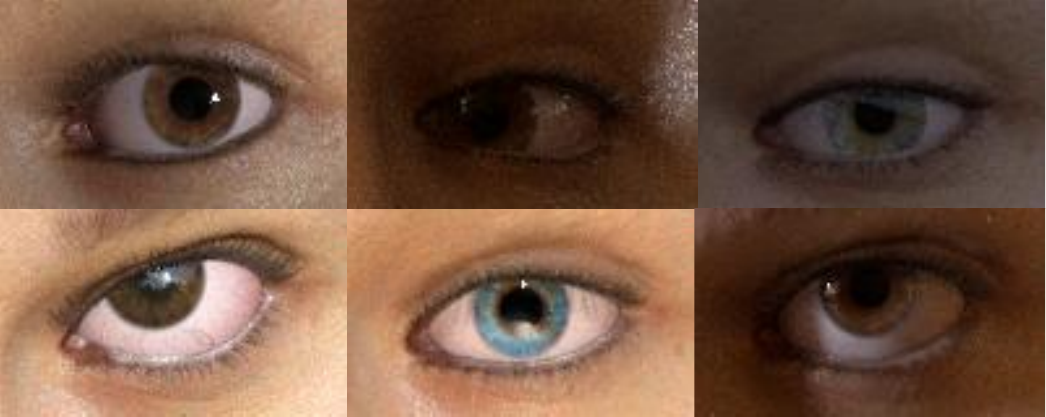}
&
\includegraphics[scale=0.5]{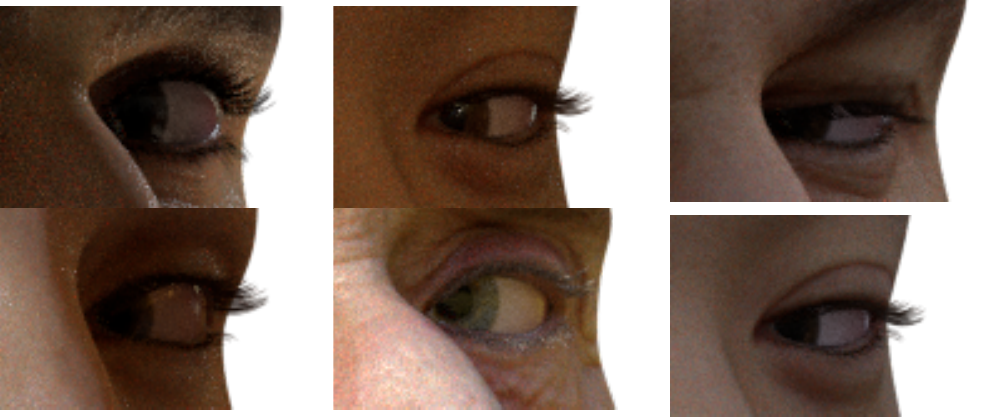}\\
(a) & (b) & (c)
\end{tabular}
\vspace{-1em}
\caption{Example eye images from our high-quality synthetic gaze dataset with (a) large positive head pose yaw angles,  (b) near-zero head poses, and (c) large negative head pose yaw angles.
	      Note how eye images with different head poses have significantly different appearances.}
\label{fig:EyeImages}
\vspace{-1em}
\end{figure*}

\section{Method}

\subsection{Head Pose Dependence}

The goal of unconstrained gaze tracking is to estimate the angle between the visual axis of the eye and that of a remotely placed stationary camera that observes the subject, all while allowing the subject to move their head freely. 
Head pose is not strictly required to compute gaze direction and it can be computed purely based on the shape of the pupil and iris in the camera's image \cite{hansen2010eye}. 
However, for machine learning regressors trained to regress gaze direction from eye images, changing head pose introduces a number of sources of variability into the problem.

First, for the same gaze angles, head rotation causes significant change in the appearance of the observed eye images. 
For example, Figure ~\ref{fig:EyeImages} shows various eye images with nearly identical gaze angles close to zero, but with large variation in head pose. 
In all these images, while the shape of the pupil and iris region is similar (almost circular because of the zero gaze angle), differences in head pose cause (a) the pupil and iris region to be occluded differently, and (b) significant differences in the appearance of the sclera and skin regions, surrounding the pupil.
Nevertheless, when presented with sufficient and variable training examples a CNN model with adequate capacity may learn to disregard these appearance changes as non-contributing factors to determining gaze direction.

\begin{figure}[b!]
\centering
\vspace{-1em}
\begin{tabular}{cc}
\includegraphics[scale=0.125]{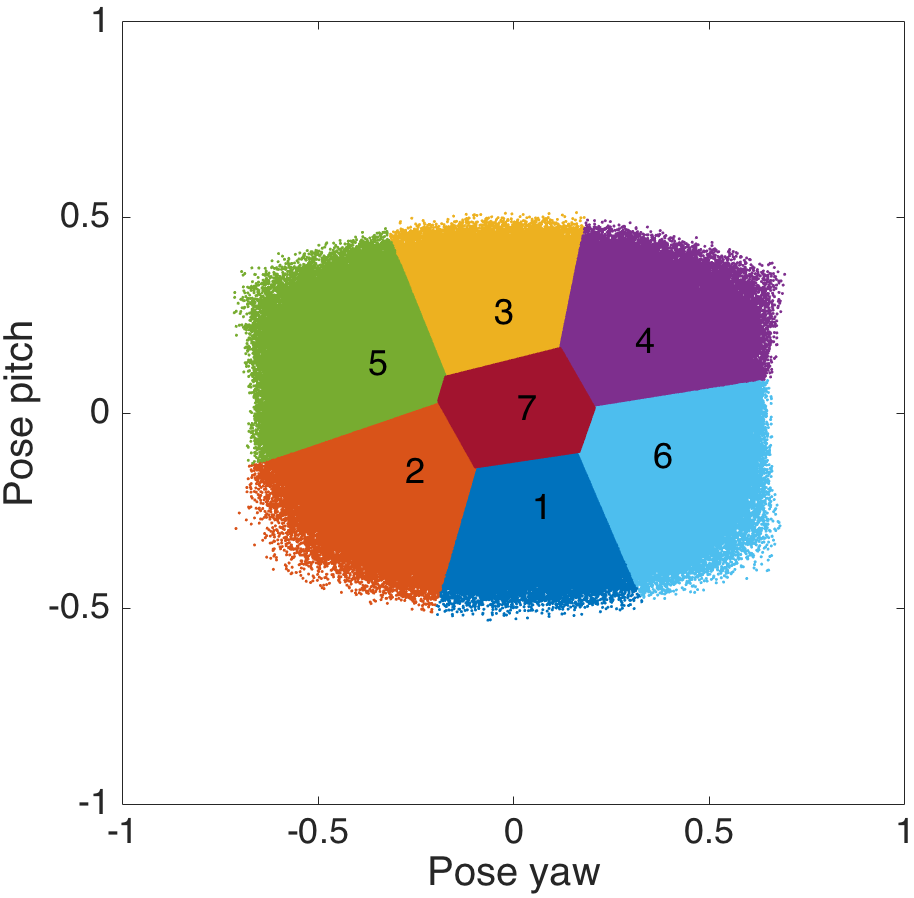}
& 
\includegraphics[scale=0.26]{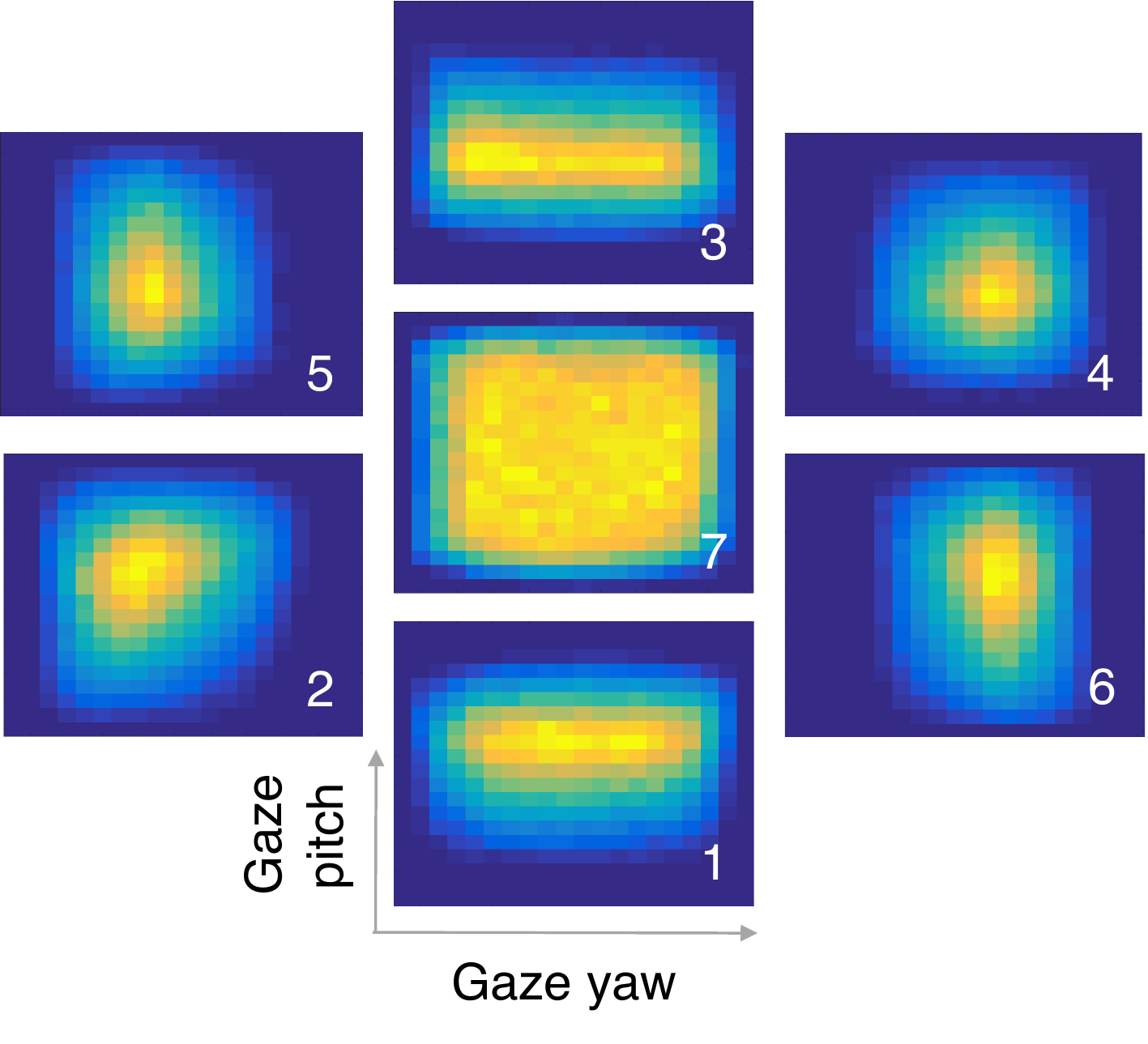}\\
(a) & (b)
\end{tabular}
\vspace{-1em}
\caption{This figure contains (a) 7 head pose clusters for our synthetic dataset, and (b) the 2D distributions of the gaze angles for these clusters.}
\label{fig:synth_clusters}
\end{figure}

The second effect of variable head pose, which is more subtle to understand and harder to solve is that with variable head pose, the prior distribution of gaze angles that a machine learning regressor needs to predict changes.
To understand this effect, we create a large synthetic dataset in-house by rendering 1M eye images using the SynthesEyes~\cite{wood2015rendering} computer graphics model. 
We render eye images with uniformly distributed $[-60,60]^{\circ}$ head pitch (up/down) and yaw (left/right) angles, and uniformly distributed eye-in-head rotation pitch and yaw angles of $[-25, 25]^{\circ}$ and $[-35, 35]^{\circ}$, respectively, distributed about each head pose.
We cluster head poses into various clusters (using cosine distance) shown in Figure~\ref{fig:synth_clusters}(a).
The distributions of gaze pitch and yaw angles for different head pose clusters are shown in Figure~\ref{fig:synth_clusters}(b).
Observe the distinctly different distributions of gaze angles for the various head pose clusters.

The reasons for this phenomenon are two-fold. 
First, as the head rotates, the eyeballs rotate in synchrony with it and gaze angles for a particular head orientation tend to be distributed roughly centrally about the instantaneous head pose. 
Hence, the mean of the distribution of gaze angles is a function of the observed head pose. 
Second, head rotation causes the shape of the prior distribution of the gaze angles to change (see Figure~\ref{fig:synth_clusters}(b)). 
To understand why this happens consider a stationary camera that observes a frontal face, (\textit{e.g.}, the cluster number 7 in Figure~\ref{fig:synth_clusters}(a)). 
For this head pose, for the entire range of motion of the eyeball, the eye, pupil and iris regions are visible to the camera and hence all gaze angles corresponding to these eyeball rotations can be regressed by the CNN. 
As the head rotates, either the eye or the pupil and iris regions, for certain eyeball rotations, become invisible to the camera. 
For example, for a head turned to the left, the left eye may not be visible to the camera. 
Additionally, even for the visible right eye, its pupil and iris may be visible in the camera image only for rotations of the eyeball to the right relative to the head. 
Hence for oblique head poses, eye-image-based CNNs are presented with eye images with limited gaze angles and hence have smaller distributions of gaze angles that they need to regress (see clusters 1-6 in Figure~\ref{fig:synth_clusters}(b)).

\begin{figure}[b!]
\centering
\begin{tabular}{cc}
\includegraphics[scale=0.125]{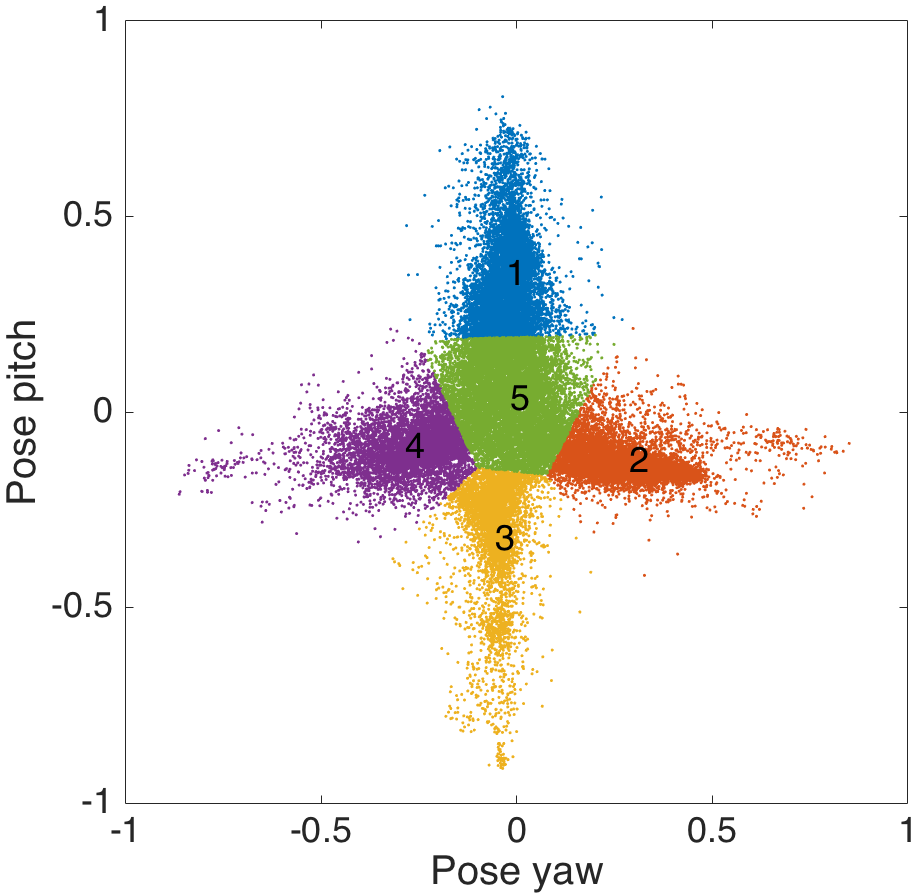}
& 
\includegraphics[scale=0.26]{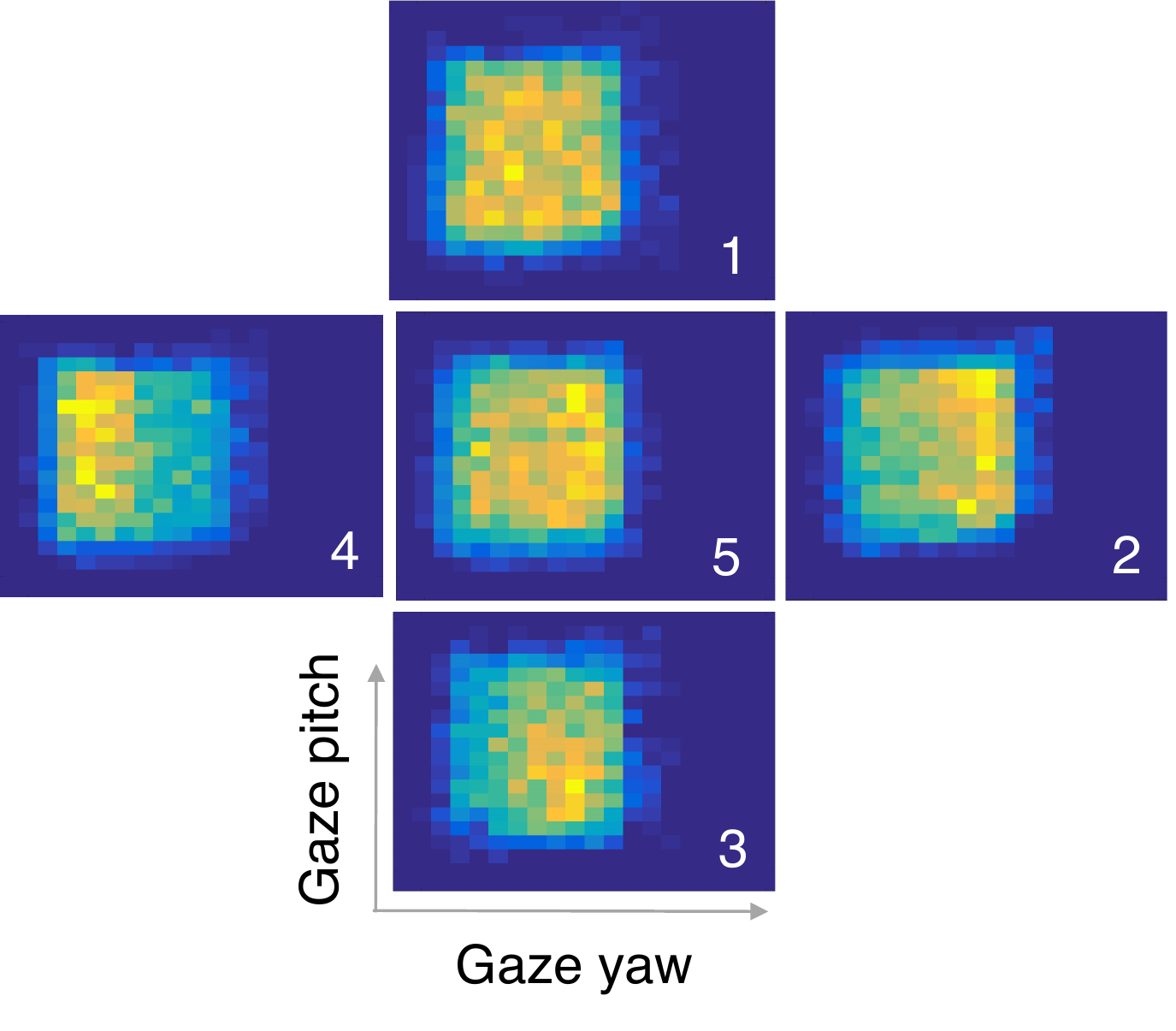}\\
(a) & (b)
\end{tabular}
\vspace{-1em}
\caption{This figure contains (a) 5 head pose clusters for the MPIIGaze dataset, and (b) the 2D distributions of the gaze angles for these clusters.}
\label{fig:mpii_clusters}
\end{figure}

We observe a similar trend for the real-world MPII Gaze dataset~\cite{zhang2015appearance} as well on clustering it into 5 head pose clusters in Figure~\ref{fig:mpii_clusters}.
Note that this dataset was collected while subjects viewed their laptop screens and hence contains only a narrow range of head poses angles.
This explains the noticeable, but less prominent differences in the distributions of gaze angles for the various head pose clusters of this dataset (Figure~\ref{fig:mpii_clusters}(b)) versus the synthetic dataset (Figure~\ref{fig:synth_clusters}(b)).

These observations about the effect of head pose on the problem of unconstrained appearance-based gaze estimation via machine learning methods, motivate the need to incorporate greater specificity in their design for different head poses. In the following section, we describe one such novel CNN architecture that we propose to tackle this problem. 



\subsection{Branched Gaze Network}
A naive approach to tackling this problem of variable head pose is to train multiple gaze networks, each for a given head pose. 
However, this is infeasible and would result in a very large number of classifiers.
An alternative solution is to train a network specific to a group of head poses with small variations. 
Hence, we cluster the head poses in our training data into $K$ groups using k-means clustering and cosine distance. 
The value of $K$ is chosen empirically via cross-validation.

One obvious solution is to train multiple gaze networks, one for each head pose cluster. 
Zhang \etal applied this approach in~\cite{zhang2015appearance, zhang2017mpiigaze} and observed a decrease in accuracy with multiple gaze networks.
Using different CNNs for each head pose cluster has the drawback of data scarcity during training.
For $K$ clusters, the number of training samples for each network decreases to roughly $1/K^\mathrm{th}$ of the original size of the training dataset. 

\begin{figure}[htp!]
      \centering
      \vspace{-1em}
      \includegraphics[width=0.48\textwidth]{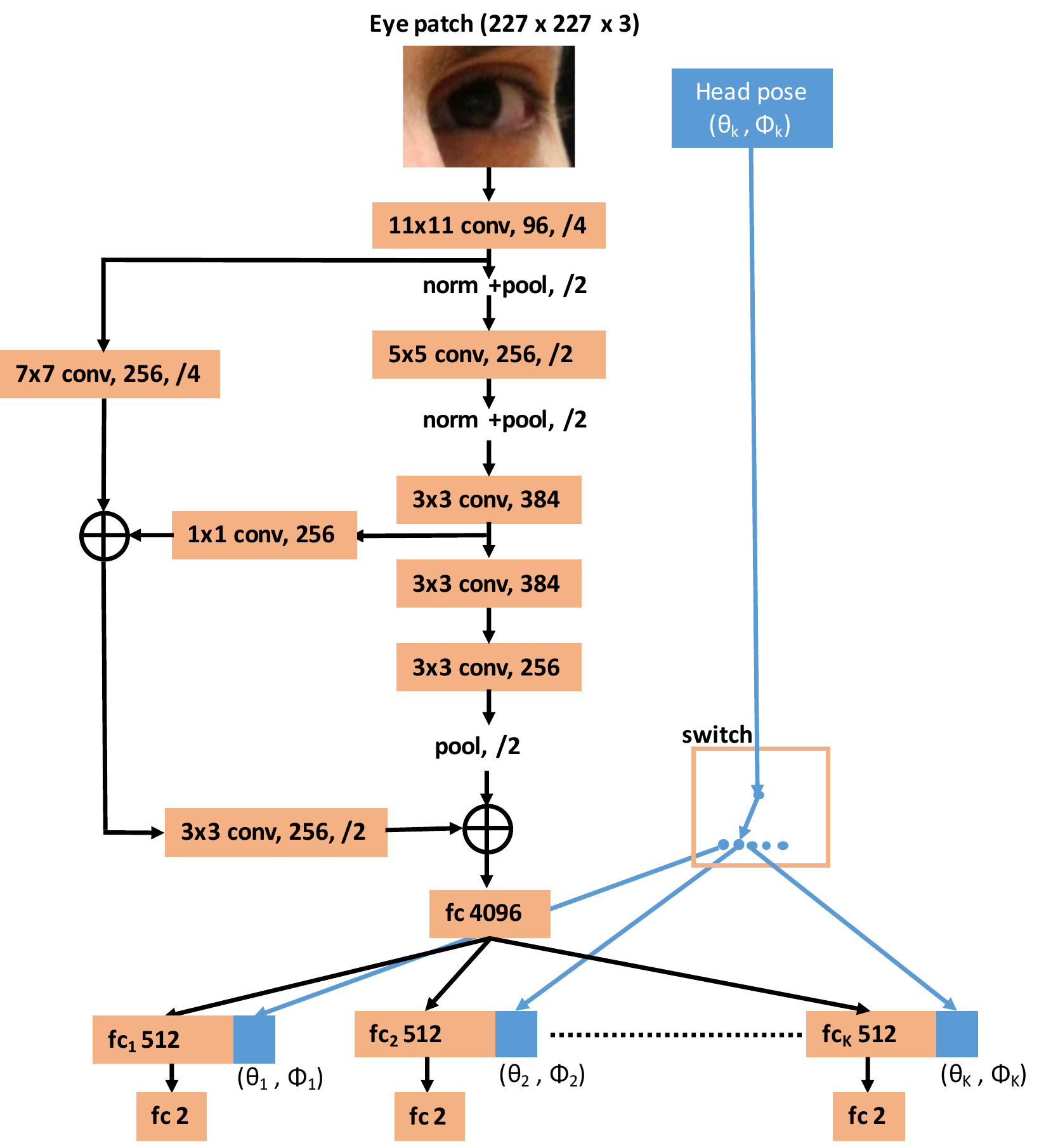}
      \vspace{-1em}
      \caption{The proposed CNN architecture for estimating gaze angles from the image of an eye and head pose. The annotations in each box represent in order -- the filter kernel sizes, the CNN layer type, the number of output feature maps, and the stride.}
      \label{fig:architecture}
      \vspace{-1em}
\end{figure}

To overcome this issue, we instead use weight-sharing among the different gaze networks for head pose clusters. 
We use a modified version of the Alexnet~\cite{krizhevsky2012imagenet} CNN architecture, as shown in Figure~\ref{fig:architecture}. 
It consists of 5 convolutional layers (conv1-conv5) followed by a max pooling layer (pool5) and a fully connected layer (fc6).
We also incorporate residual skip connections for the classifier to have easy access to the lower level features of the network.
We share the weights of the lower layers (conv1 to fc6) across all the head pose clusters, but add multiple fully connected (fc7 and fc8)  layers on top of the fc6 layer, one pair for each head pose cluster.
Similar to Zhang et al.~\cite{zhang2015appearance}, we concatenate head pitch and yaw angles with the fc7 layers of all head pose clusters.
The final output layers (fc8) produce the pitch and yaw gaze angles (in radians). 
We use the rectified Linear Unit (ReLU) as the activation function after all convolution and fully connected layers.

The inputs to our network are an eye image, head pitch and yaw angles and head pose cluster ID. 
During training, only the training cases that belong to a particular cluster update the weights of their corresponding fc7-fc8 layers, but all training samples from all clusters update the conv1-fc6 layers, which reduces the problem of data scarcity. 
Additionally, during inference, based on the known cluster ID, each input is conditionally routed through only one pair of fc7-fc8 layers corresponding to its pose cluster. This ensures no increase in computational cost per input versus an un-branched network. Thus our proposed architecture is efficient in terms of training and testing.



\subsection{Training}
We describe a number of effective techniques that we employ to train our gaze classifier, which help to improve its accuracy significantly.

\subsubsection{Object Viewpoint Estimation}
Transfer learning from CNNs trained for the task of object recognition on the large ImageNet dataset~\cite{russakovsky2015imagenet} is a widely used approach, especially when the amount of training data available for the task at hand is small.
This approach is also used exclusively for initializing all the successful gaze networks~\cite{zhang2015appearance, krafka2016eye, zhang2017s, zhang2017mpiigaze, deng2017monocular}.
A careful visualization of the activations of a network trained on ImageNet~\cite{zeiler2014visualizing} suggests that its earlier layers capture low level features that describe the location and orientation of an object, while the deeper layers capture the global characteristics of the objects and are invariant to their position and viewpoint. 
Their insensitivity to object pose makes networks trained for object recognition not ideal for initializing networks for gaze estimation. Gaze estimation fundamentally involves estimating the 3D orientation of the eyeball.

We instead transfer the learning of a network trained for general object viewpoint estimation, which is a more closely related task to gaze estimation than object recognition.
We initialize the weights of our gaze network with the pre-trained weights of the ``Render for CNN"~\cite{su2015render} (RFC) network, which shares architectural elements with our gaze network. 
The RFC network was designed to estimate the 3D orientation of 20 categories of objects, including cars, airplanes, chairs, etc.
Additionally, it was trained using a large corpus (2.4M) of synthetically rendered images of different object categories.
We show through multiple experiments that since RFC is designed to distinguish between subtle changes in the appearance of objects caused by their rotation, it is a better choice as a pre-trained network for gaze estimation.

\subsubsection{High-quality Synthetic Data}

We create a dataset of 1M synthetic eye images by rendering left eye images of size 134$\times$80 pixels using advanced graphics techniques and ray tracing with the SynthesEyes model~\cite{wood2015rendering}.
It contains high-quality 3D scans of 10 human head models (5 female and 5 male from different age groups and ethnicities) and detached eyeball models, which can be posed independently of the head models.
Additionally, the subjects' eyelids can be posed in synchrony with the up/down movement of the eyeball.
The images contain wide variations in illumination, which is achieved by image-based relighting with different HDR environment maps.
Some example images from this synthetic dataset are shown in Figure \ref{fig:EyeImages}.
These images are of much higher quality in terms of photo-realism than the popular real-time rendered UnityEyes dataset~\cite{wood2016learning}.
Our work is the first to create and use such a large dataset of high-fidelity rendered eye images.

For all experiments with synthetic data, we transfer learn from the RFC network only.
For the real-world datasets we perform an additional step of training on the 1M synthetic data first and then fine-tuning with the real data.
Similar to previous work~\cite{wood2015rendering, zhang2017mpiigaze}, we too observe that to effectively transfer learn form synthetic data, it is important that both datasets have the same distribution of head pose and gaze angles.
Hence we target the synthetic dataset to match the distribution of the real dataset.

\section{Experiments}

\subsection{Synthetic Data}

\subsubsection{Network Initialization}
\label{section:NetworkInitialation}

We evaluate the performance of our gaze network with different weight initializations. 
For all experiments, we use 800K images of 8 subjects from the the 1M synthetic dataset for training and 200K images of two randomly selected subjects (female 2 and male 4) for testing.
The inputs to all networks are normalized eye images and head pitch and yaw angles (radians), which are computed with the normalization procedure outlined in~\cite{zhang2015appearance}.
In this experiment, all networks are un-branched, meaning they contain one set of fc layers for all head pose clusters.
We initialize our gaze network with (a) random weights, (b) weights from AlexNet~\cite{krizhevsky2012imagenet} trained for object recognition on ImageNet, and (c) with weights from the RFC~\cite{su2015render} network trained for object viewpoint estimation.
Additionally, we also implement a gaze network based on the more complex ResNet-50 network~\cite{he2016deep} with head pose concatenated with its pool5 layer and initialize it with weights from ImageNet.

%
%

\begin{table}[t]
	\centering
    \caption{The angular gaze errors for different network architectures (AlexNet and ResNet-50) initialized with different pre-trained networks.}
    \vspace{-1em}
    \label{table:CNN_architecture}
    \begin{tabular}{lccc}
		\toprule
		Network & Weights Initialization & Error ($^{\circ}$) 	\\
		\midrule	
		Ours & ImageNet~\cite{krizhevsky2012imagenet} & 5.03 \\
		Ours & RFC~\cite{su2015render} & \textbf {4.40} \\
		ResNet-50 & ImageNet~\cite{he2016deep} & 5.07 \\
        	\bottomrule
	    \vspace{-1em}
    \end{tabular}
    \vspace{-1em}
\end{table}

The angular errors for gaze estimation with these networks are shown in Table~\ref{table:CNN_architecture}.
Our gaze network when initialized with random weights fails to converge, but all other networks converge. 
When initialized with weights from RFC, our network has the lowest gaze error of $4.4^{\circ}$, which is considerably smaller than initializing from ImageNet. 
Interestingly, our AlexNet-style gaze network, despite being much less complex than the ResNet-50-based network, when initialized with RFC, also performs better than it.
This shows the importance of proper initialization for training gaze networks and how transferring knowledge from the more closely related task of object viewpoint estimation versus object recognition can help it to converge to a better solution.

\subsubsection{Effect of Head Pose}
\begin{table}[b]
	\centering
    \caption{The effect of head pose on the accuracy of gaze estimation of our network. We compare the gaze errors ($^{\circ}$) for networks (a) with and without head pose as an input and (b) with and without branching of the fully connected fc7-8 layers for the different head pose clusters.}
    \vspace{-1em}
    \label{table:Inputs_outputs}
    \begin{tabular}{lcccc}
		\toprule
        Input  & fc7-8 & Experiment I & Experiment II	\\
 	\midrule
	Eye & single & 5.66 & 5.76 \\
	Eye & branched& 5.18 & 4.75\\
	\midrule
	Eye, pose  & single & 4.40 & 3.91 \\
	Eye, pose  & branched & \textbf{4.15} & \textbf{3.68}\\
	\bottomrule
    \end{tabular}
\end{table}

\label{section:HeadPose}

\begin{table*}[!t]
	\centering
    \caption{The individual gaze errors ($^{\circ}$) of each of the head pose clusters of the synthetic dataset shown in Figure~\ref{fig:synth_clusters}. These errors are
    		for our gaze network without head-pose-dependent branching of the fc7-8 layers (second row) and with branching (third row).}
    \vspace{-1em}
    \label{table:cluster_accuracy}
    \begin{tabular}{lccccccc|cc}
	\toprule
	CNN Architecture & Cluster 1 &  Cluster 2 &Cluster 3 & Cluster 4 & Cluster 5 & Cluster 6 & Cluster 7 & Overall \\
 	\midrule
	single fc7-8 & 4.13 & 4.43 & 4.02 & 4.59 & 4.33 & 4.02 & 4.61 & 4.40 \\
	\midrule
	branched fc7-8 & \textbf{4.09} & \textbf{4.19} & \textbf{3.92} & \textbf{4.08} & \textbf{4.31} & \textbf{3.92} & \textbf{4.61} & \textbf{4.15} \\
	\bottomrule
    \end{tabular}
    \vspace{-1em}
\end{table*}

We investigate the effect of head pose on our gaze network. 
We evaluate the performance of our gaze network (a) with and without head pose as input and (b) with and without late branching of the network for different head pose clusters.
We use the same experimental protocol (Experiment I) described in Section~\ref{section:NetworkInitialation}.
In addition, we perform a leave-one-subject-out experiment (Experiment II) where we train with ~900K images from 9 subjects and test on the 100K images of the one held-out subject. 
We repeat this experiment 10 times for all subjects and average the gaze errors.
 
The results are shown in Table~\ref{table:Inputs_outputs}.
We observe that providing head pose as input to our network, in addition to the eye image, has a significant positive impact on its accuracy.
Additionally, branching the high-level fc7-8 layers for different head pose clusters results in improved accuracy of gaze estimation.
The positive effect of branching is particularly pronounced when the head yaw and pitch angles are not provided as inputs to the gaze network.
These observations are consistent for both Experiment I and Experiment II.

To better understand how our gaze network behaves with and without head pose-dependent branching, we compute the gaze errors (Table~\ref{table:Inputs_outputs}) for the 7 individual head pose clusters as shown in Figure~\ref{fig:synth_clusters} for Experiment I.
Obsreve that with the branched architecture (row 3 of Table~\ref{table:cluster_accuracy}), the accuracy of all head pose clusters either remains the same or improves versus the un-branched architecture (row 2 of Table~\ref{table:cluster_accuracy}).
Clusters 2 and 4, which contain oblique head poses, benefit the most from the branched network architecture, while the central cluster (number 7), which contains head pose angles close to zero remains unchanged.
Plausible explanations for this observation are that since head pose is close to null and the gaze angles are less strongly correlated to the head pose for this cluster, head pose has minimal impact on it.

Overall, these results confirm the presence of correlations between oblique head poses and gaze angles, which, given sufficient data can be exploited to improve the accuracy of appearance-based gaze estimation via machine leaning regressors.
Our branched gaze network design is one such proposed architecture that helps to improve the accuracy of gaze estimation by effectively using head pose information, while being efficient in terms of training and inference.

\subsubsection{Number of subjects}

On comparing the results of Experiment I and Experiment II in Table~\ref{table:Inputs_outputs}, one can also conclude that having greater subject variability in the training data helps to improve the accuracy of our gaze network.
We perform an additional experiment where we train our network with 80\% of the data from all 10 subjects in the synthetic dataset and test on the remaining 20\%.
This produces an an accuracy of $1.99^{\circ}$ and represents the accuracy of our gaze network for the person-specific training scenario.

\subsection{Real-world Data}


We evaluate the performance of our gaze network and the efficacy of our training procedure, which involves transfer learning from RFC and the 1M synthetic dataset on the real-world benchmark Columbia~\cite{smith2013gaze} and MPIIGaze~\cite{zhang2015appearance} datasets. 
For these datasets, we mirror the right eye images about the vertical axis and reverse the signs of their gaze and head yaw angles to process all images with one gaze network as was done in~\cite{zhang2015appearance}. 

\begin{figure}[t]
\centering
\begin{tabular}{c c}
\includegraphics[scale=0.36]{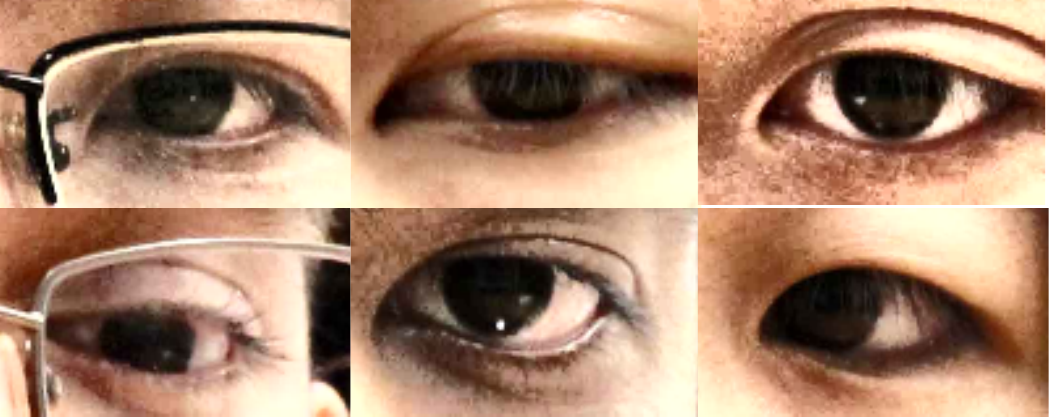}
&
\includegraphics[scale=0.36]{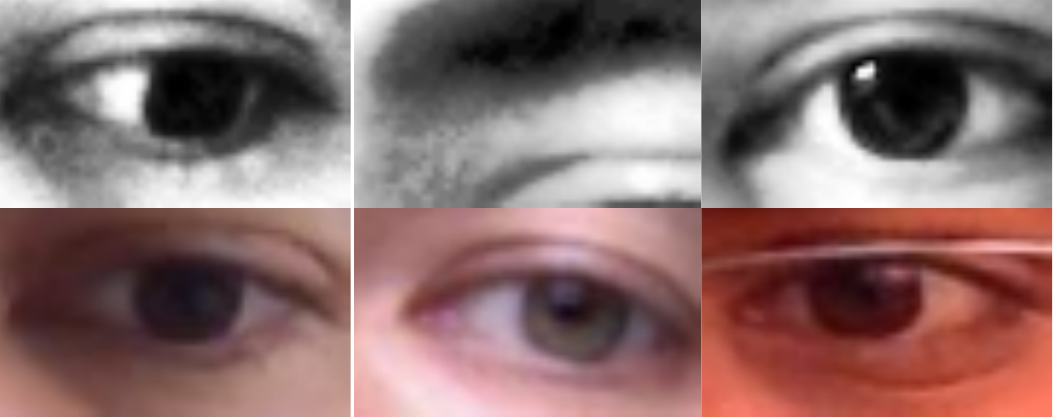}\\
(a) & (b)
\end{tabular}
\vspace{-1em}
\caption{Examples of normalized eye images from (a) the Columbia gaze dataset, and (b) the MPIIGaze dataset's ``Evaluation" set (top) and ``Annotation" set (bottom). Note the erroneous image (top-center) from the MPIIGaze dataset.}
\label{fig:EyeImagesReal}
\vspace{-1em}
\end{figure}

\begin{table}[b]
	\centering
    \caption{The gaze errors ($^{\circ}$) of our gaze network and its comparison to the current state-of-the-art~\cite{wood20163d} on the Columbia dataset~\cite{smith2013gaze}. 
    		The second row contains errors for only 34 subjects without eyeglasses, while the third row contains the errors of all 56 subjects with/without eyeglasses.}
    \vspace{-1em}
    \label{table:Columbia}
    \begin{tabular}{lcccc}
	\toprule
	Subjects & Wood~\cite{wood20163d} & \multicolumn{2}{c}{Ours branched} \\
	&  & RFC & 1M Synthetic \\	
	\midrule
	No glasses &7.54 & 6.26 & \textbf{5.58} \\
	\midrule
    	All & - & 6.68 & \textbf{5.65} \\
	\bottomrule
    \end{tabular}
\end{table}

\subsubsection{Columbia Dataset}

\begin{table*}[!t]
	\centering
    \caption{A comparison of the performance of our proposed algorithm against the comparable state-of-the-art algorithm on the MPIIGaze dataset~\cite{zhang2015appearance}. 
    		The table contains the average angular errors ($^{\circ}$) across 15 repetitions of the leave-one-subject-out experiment on the MPIIGaze dataset.
		}
    \vspace{-1em}
    \label{table:MPIIGaze}
    \begin{tabular}{cc|ccccccc}
	\toprule
	& Zhang~\cite{zhang2015appearance} & & \multicolumn{3}{c}{Ours single} & \multicolumn{3}{c}{Ours Branched}  \\
	 & 12K Synth~\cite{wood2015rendering} & & ImageNet & RFC & 1M Synth & ImageNet & RFC & 1M Synth\\	
	\midrule
	MPIIGaze & 5.5 & Evaluation & 5.88 &  5.56 & \textbf{5.38} &5.77 & 5.49 & 5.48\\
	\midrule
    	MPIIGaze+ & 5.4 & Annotation & 5.88 & 5.38 & \textbf{5.30} & 5.63 & 5.46 & 5.42 \\
	\bottomrule
    \end{tabular}
    \vspace{-1em}
\end{table*}

The Columbia gaze dataset~\cite{smith2013gaze} contains 5,880 images of 56 subjects (22 with eyeglasses and 34 without), captured indoors in a laboratory from 5 viewpoints with head pose yaw angles ranging from $-30^{\circ}$ and $30^{\circ}$ in steps of $15^{\circ}$. 
Additionally, for each viewpoint, images for 21 different gaze directions are present. 
We extract normalized color eye images of size 134$\times$80 pixels and compute the normalized head pose and gaze directions, using the procedure of~\cite{sugano2014learning}.
We also apply histogram equalization to the Y channel of the YCbCr representation of the images and convert them back to RGB.
Example normalized images are shown in Figure~\ref{fig:EyeImagesReal}(a).

We divide the 56 subjects into three partitions and for each partition, we train with two-thirds of the subjects and test with the remaining one-third.
We repeat this thee-way cross-validation-across-subjects procedure three times and average the gaze errors for all subjects.
We cluster the Columbia data into 5 head pose clusters centered about the five discrete head yaw angles present in this dataset.
We compute the performance of our gaze network for initialization with (a) RFC only and (b) with RFC followed by fine tuning with a targeted subset of the 1M synthetic dataset, sampled to match the gaze and head pose distribution of the Columbia dataset.
Wood \etal~\cite{wood20163d} report the state-of-the-art error of $7.54^{\circ}$ with their method of morphable eye-region matching on a randomly selected set of 680 images of 20 subjects without eyeglasses from this dataset. 
To compare against their results, we also repeat our three-fold cross validation experiment on the subset of 34 subjects without eyeglasses.

The results are shown in Table~\ref{table:Columbia}.
Our gaze network produces the best known accuracy on this dataset with a gaze error of $5.58^{\circ}$ for images without eyewear and $5.65^{\circ}$ for all images.
Thus there is only a marginal decrease in accuracy on including images with eyeware.
Both versions of our gaze network, trained with and without the synthetic data, outperform Wood \etal's algorithm.
Furthermore, the additional step of  pre-training with synthetic data and then fine tuning with the real-world data (column 3) helps to improve the accuracy of our gaze network by nearly $1^{\circ}$.

%

\subsubsection{MPIIGaze Dataset}

The MPIIGaze dataset~\cite{zhang2015appearance} contains images of 15 subjects (with/without eyeglasses) captured while they were viewing their laptop screens using the laptop's front-facing camera. 
The images contain widely varying illuminations.
The dataset contains an ``Evaluation" set of 45K normalized grayscale eye images (3K per subject) of size 60$\times$36 pixels with their accompanying normalized head pose and gaze angles (top row of Figure~\ref{fig:EyeImagesReal}(b)).
For normalizing the images in the ``Evaluation" set, the authors use an automatic facial fiducial point detection algorithm, which is erroneous 5\% of the time. 
This in turn produces erroneous normalized images that often do not even contain an eye (\textit{e.g.}, top-center image of Figure~\ref{fig:EyeImagesReal}(b)).
In our analysis, we include these erroneous images as well and did not discard them.
In addition to the ``Evaluation" set, the authors also provide an ``Annotation" Set of 10,848 images, which contain manually annotated facial fiducial points.
We normalize the images in the ``Annotation" set and evaluate the performance of our gaze network on it as well.


We follow the same leave-one-subject-out experimental protocol that was previously employed in~\cite{zhang2015appearance, zhang2017mpiigaze}. 
We train with images from 14 subjects, test on images of the left-out subject and repeat this procedure 15 times. 
We report the average errors across all repetitions.
We evaluate the accuracy of both the un-branched and branched versions of our gaze network on the MPIIGaze dataset.
For the branched network, we cluster the head poses into 5 clusters (Figure~\ref{fig:mpii_clusters}(a)) using k-means clustering and cosine distance.
Furthermore, we evaluate initialization of the weights of our gaze network with the RFC network with and without further training on the 1M synthetic dataset.


Recently, Zhang \etal~\cite{zhang2017mpiigaze} report the state-of-the-art gaze error of $5.5^{\circ}$ on the ``Evaluation" set of the MPIIGaze dataset with erroneous facial fiducial point annotations included in it.
They also report a gaze error of $5.4^{\circ}$ for another version of the ``Evaluation" subset called (MPIIGaze+), wherein, they manually annotate the facial fiducial points.
Note that the MPIIGaze+ dataset is not publicly available. 
The backbone of their gaze network is the VGG network~\cite{simonyan2014very}, which is 10$\times$ slower than our gaze network. 
Additionally, similarly to us, they initialize their gaze network with weights from ImageNet, first train with a targeted subset of the 12K synthetic images generated by Wood \etal~\cite{wood2015rendering}, and later fine-tune on the MPIIGaze dataset for cross-subject evaluation.

%


\begin{table*}[t!]
	\centering
    \caption{The cross-dataset angular errors ($^{\circ}$) of various state-of-the-art gaze networks and ours on the ``Evaluation" set of the MPIIGaze dataset.}
    \vspace{-1em}
    \label{table:crossDataset}
    \begin{tabular}{lcccccc}
	\toprule
	&Dataset & Network & Initialization & Domain Adaptation & Error ($^{\circ}$) \\
	\midrule
	Shrivastava~\cite{shrivastava2017learning}  & UnityEyes 1M & LetNet~\cite{zhang2015appearance} & Random & GAN & 7.80 \\
	Zhang~\cite{zhang2017mpiigaze} & UT Multiview & VGG & ImageNet &  targeting & 8.0\\
	Zhang~\cite{zhang2017mpiigaze} & SynthesEyes 12K & VGG & ImageNet & targeting &\textbf{7.3}\\
	\midrule
	Ours &  SynthesEyes 1M & AlexNet & ImageNet & none & 10.19 \\	
	Ours &  SynthesEyes 1M & AlexNet & RFC &none & 8.72 \\	
	Ours &  SynthesEyes 1M & AlexNet & RFC & targeting & 7.74 \\	
	\bottomrule
    \end{tabular}
    \vspace{-1em}
\end{table*}

The results of these experiments are presented in Table~\ref{table:MPIIGaze}.
We observe that the un-branched version of our gaze network when initialized with weights from pre-trained RFC network and the synthetic data produces the lowest gaze errors for both the ``Evaluation" and ``Annotation" subsets of the MPIIGaze dataset of $5.4^{\circ}$ and $5.3^{\circ}$, respectively.
These are the lowest reported gaze errors on this dataset.
Furthermore, our gaze network outperforms the state-of-the-art network of Zhang \etal~\cite{zhang2017mpiigaze}, both for automatically and manually annotated facial fiducial points, despite the fact our network is $10\times$ faster and significantly smaller than theirs.
Proper initialization of the weights of our gaze network from the more closely related task of object viewpoint estimation performed by the RFC network contribute considerably towards improving its accuracy.
We observe a further improvement by employing the synthetic data for pre-training the network.

Similar to what Zhang \etal~\cite{zhang2017mpiigaze} report, for the MPIIGaze dataset, we too observe a decrease in accuracy with head-pose-dependent branching. 
As Zhang \etal~\cite{zhang2017mpiigaze} also note, it is not surprising that the effect of head pose is minimal for this dataset as it mostly contains near-frontal head poses.
However, the results of our previous simulation analyses with the larger, denser and much more variable 1M synthetic dataset confirm that in the presence of large head pose variations the branched CNN architecture is beneficial to improve the accuracy of gaze estimation.


\subsubsection{Cross-dataset Experiment}

Lastly, we evaluate the performance of our gaze network for the cross-dataset evaluation protocol.
We train our gaze network (without branching) on the 1M synthetic dataset with and without targeting it to match the distribution of the MPIIGaze dataset.
We test its performance on real-world images from the ``Evaluation" set of the MPIIGaze dataset.
We also evaluate the performance of various weight initializations: ImageNet and RFC.

Table~\ref{table:crossDataset} shows the gaze errors for the various configurations of our gaze network and for other gaze estimation algorithms.
Shrivastava \etal's~\cite{shrivastava2017learning} use 1M images from the low quality UnityEyes~\cite{wood2016learning} dataset for training, but they adapt the appearance of the rendered images to match that of the real images using generative adversarial networks. For gaze estimation they use the original LeNet-style network proposed by Zhang \etal~\cite{zhang2015appearance}.
Zhang \etal~\cite{zhang2017mpiigaze}, recently experiment with the UT MultiView~\cite{sugano2014learning} and the 12K SynthesEyes~\cite{wood2015rendering} datasets and subsample them to target the distributions of the synthetic datasets to that of the real MPIIGaze dataset. They use the more complex VGG network for gaze estimation and achieve the lowest gaze errors of all others.

Our algorithm is competitive to the other competing algorithms in terms of gaze error.
Furthermore, as we observe in the other experiments, here too initializing the weights with the RFC is the major contributing factor in improving the accuracy of our network.
Targeting of the synthetic dataset also helps to improve its accuracy.


%


\section{Conclusions}
Unconstrained gaze estimation using off-the-shelf cameras that observe subjects from a remote location is a non-trivial problem as it requires handling many different head poses.
We systematically study the affects of head pose on the problem of gaze estimation via machine learning regressors.
We show that head pose changes the appearance of eye images along with the prior distribution of gaze angles that the machine learning regressor must predict.
To address this, we cluster head poses into various clusters and design a branched gaze network, which contains shared lower-level layers for all head pose clusters, but multiple cluster-specific higher-level layers.
This design is efficient in terms of training and inference.
We show via a large simulation study containing 1M synthetic images that in the presence of large head poses, such a branched network is beneficial to improve the accuracy of gaze estimation.

Lastly, we propose new effective procedures for training gaze networks including transfer learning from the more closely related task of general object viewpoint estimation versus object recognition, and from a large synthetic dataset of high-quality rendered images. We show that these training procedures help our light-weight AlexNet-style gaze network achieve best-in-class accuracy on two real-world datasets, despite being significantly ($10\times$) faster and smaller than its directly competing state-of-the-art algorithm.

%
%
%
%

{\small
\bibliographystyle{ieee}
\bibliography{gaze_for_review_cvprw}
}

\end{document}